\documentclass[letterpaper, 10 pt, conference]{ieeeconf}  

\IEEEoverridecommandlockouts                              
\usepackage[T1]{fontenc}
\usepackage[utf8]{inputenc}
\usepackage{array}
\usepackage{wrapfig}
\usepackage{multirow}
\usepackage{tabularx}
\usepackage{booktabs}
\usepackage{hyperref}
\usepackage{cite}
\usepackage{ulem}
\usepackage{graphicx}
\usepackage[ruled,vlined]{algorithm2e}
\usepackage{makecell}

\usepackage{algorithmic}
\usepackage{setspace}
\usepackage{textcomp}
\usepackage{xcolor}
\usepackage{amsmath,amssymb,amsfonts}

\usepackage{booktabs}
\usepackage{multirow}

\usepackage{subfigure}

\newcommand{\sqkm}{$\text{km}^2$}
\newcommand{\ourMethod}{SCoPP}

\title{\LARGE \bf
Scalable Coverage Path Planning of Multi-Robot Teams for Monitoring Non-Convex Areas
}

\author{Leighton Collins$^{1}$, Payam Ghassemi$^{1}$, Ehsan T. Esfahani$^{1}$, \\ David Doermann$^2$, Karthik Dantu$^2$, and Souma Chowdhury$^{1\dagger}$

\thanks{$^{1}$ L.C., P.G., S.C. and E.E. are with the Department of Mechanical and Aerospace Engineering, University at Buffalo, Buffalo,
NY, 14260 USA.}%
\thanks{$^{2}$ K.D. and D.D. are with the Department of Computer Science and Engineering, University at Buffalo, Buffalo,
NY, 14260 USA.}%
\thanks{$^\dagger$ Corresponding Author, 
{\tt\small soumacho@buffalo.edu}}
\thanks{* This work was supported by the DARPA award HR00111920030 and NSF award IIS-1927462. Any opinions, findings, conclusions, or recommendations expressed in this paper are those of the authors and do not necessarily reflect the views of the DARPA or the NSF.}
\thanks{\copyright 2021 IEEE. Personal use of this material is permitted.  Permission from IEEE must be obtained for all other uses, in any current or future media, including reprinting/republishing this material for advertising or promotional purposes, creating new collective works, for resale or redistribution to servers or lists, or reuse of any copyrighted component of this work in other works.}
}

\begin{document}

\maketitle

\begin{abstract}
This paper presents a novel multi-robot coverage path planning (CPP) algorithm - aka \ourMethod{} - that provides a  time-efficient solution, with workload balanced plans for each robot in a multi-robot system, based on their initial states. This algorithm accounts for discontinuities (e.g., no-fly zones) in a specified area of interest, and provides an optimized ordered list of way-points per robot using a discrete, computationally efficient, nearest neighbor path planning algorithm. This algorithm involves five main stages, which include the transformation of the user's input as a set of vertices in geographical coordinates, discretization, load-balanced partitioning, auctioning of conflict cells in a discretized space, and a path planning procedure. To evaluate the effectiveness of the primary algorithm, a multi-unmanned aerial vehicle (UAV) post-flood assessment application is considered, and the performance of the algorithm is tested on three test maps of varying sizes. Additionally, our method is compared with a state-of-the-art method created by Guasella et al. Further analyses on scalability and computational time of \ourMethod{} are conducted. The results show that \ourMethod{} is superior in terms of mission completion time; its computing time is found to be under 2 mins for a large map covered by a 150-robot team, thereby demonstrating its computationally scalability. 
\end{abstract}


\section{INTRODUCTION} \label{section:introduction}
Coverage path planning (CPP) plays a crucial role in applications such as search and rescue~\cite{kohlbrecher2014towards}, disaster response~\cite{ghassemi2019decmrta}, environmental monitoring~\cite{hodgson2016precision}, \cite{vermeulen2013unmanned}, and precision agriculture \cite{albani2017monitoring}. These applications may include large areas, and require that the designated area be scanned as quickly as possible to minimize the total time taken to locate the target, or capture as much data as possible from an area of interest (AoI) for further action (such as rescue). For applications such as post-disaster assessment and border control, it is essential to have fast and immediate solutions. Typical solutions to the proposed coverage path planning problem can be obtained through the use of available resources such as satellite imagery due to its low cost and readily accessible information. However, satellite imagery is limited in its spatial and temporal capabilities~\cite{al2013major}, drastically reducing its ability to provide data at any given time. An alternative solution is utilizing Unmanned Aerial Vehicles (UAVs) due to their accessibility and low-cost. While single-robot solutions~\cite{yu2019coverage} have been proposed to tackle coverage problems, swarm or multi-UAV systems have seen much usage in tasks akin to coverage path planning~\cite{ball2017swarm, ghassemi2020bswarm-jcise,ghassemi2018decmata}, where a quick solution for a given area is needed. A multi-robot solution typically provides greatly flexibility as well as adaptability to scenarios that are dynamic and spontaneous, such as post-disaster scenarios.  

In this paper, we propose a novel multi-robot coverage path planning algorithm with a tractable computational time suitable for planning for a large number of robots over a non-convex area with no-fly zones. This algorithm provides a set of plans for robots that complete coverage path planning with minimal required time. The applicability of the proposed method is tested on a simulated UAV swarm performing a post-disaster assessment. The remainder of this section briefly surveys the literature on multi-robot coverage path planning and area monitoring, and summarizes the contributions of this paper.



The primary contribution of this paper is a novel multi-robot coverage path planning algorithm with the following characteristics: (i) a minimized completion time by providing a balanced work load for each robot, (ii) the area of interest can be non-convex with no-fly zones, (iii) computationally amenable for coverage path planning for a large number of robots, and (iv) robots can start with different initial states or different starting point (mission with 2 or more dispatchers). 
For this purpose, we use an iterative clustering of a discretized space to partition a given area into sections to be assigned to each robot of the swarm. This partitioning method accounts for "no-fly" zones, as well as the robots states, which ensures a more even workload distribution amongst the robots, to provide a lower total time required to complete each mission. A nearest neighbor (NN) path planning algorithm is then used to reorder the list of cells assigned to each robot to optimize the path taken for each robot, in a computationally efficient manner. The algorithm proves to be scalable to hundreds of robots and large areas with a linear trend in computation time with respect to the number of robots. The secondary contribution involves testing the proposed algorithm for a post-disaster visual assessment problem by considering the current capabilities of existing unmanned aerial vehicles available on the market, as well as physical limitations due to the required image resolution, and the existing Federal Aviation Administration (FAA) regulations.

The remainder of this paper is organized as follows: The next section summarizes the existing literature for the coverage path planning and area surveillance for multi-robot systems. Section~\ref{section:methodology} presents our multi-robot coverage path planning algorithm. In section \ref{sec:case study}, a multi-UAV post-flood assessment problem is presented and a set of design of experiments that is used to evaluate the performance of 
our proposed algorithm. In section \ref{section:results} the results from the initial testing of the algorithm, as well as the case study with the given initial conditions, optimality, and viability of the algorithm as it applies to the coverage path planning problem are reviewed and evaluated. In section \ref{section:future work conclusions} the major findings as a result of creating and testing the algorithm are restated, and possible future developments that can be pursued to make the algorithm more optimal, or account for more scenarios are presented.

\section{Related Work}\label{sec:litSurvey}
\vspace{-0.2cm}

Several solutions using cellular decomposition methods have been proposed to solve the coverage path planning problem \cite{torres2016coverage, li2011coverage, guastella2019coverage}. These approaches focus on decomposing the area of interest into several partitions based on a predefined rule set, and assigning these to a set of robots. After the robots have been assigned their areas, they patrol it with a back-and-forth sweeping maneuver traveling across their area until all of it has been traversed. While efforts have been made to reduce the number of turns during the final survey maneuver \cite{torres2016coverage}, or add the ability to account for discontinuities in the environment \cite{guastella2019coverage}, these solutions do not aim to minimize the total time taken to completely survey a given area. Additionally, some solutions are tailored for optimizing specific characteristics such as energy consumption \cite{modares2017ub, di2015energy, jensen2020near} or connectivity \cite{panerati2018swarms}, but do not aim to reduce the total completion time of the mission. Similarly, while some solutions are intended for reconstruction of, or use in 3D environments \cite{englot2013three, shen2019online}, the method proposed here focuses on minimizing mission time in 2D environments.

Vehicle model selection is also crucial to the performance of the team of robots. Ahmadzadeh et al.~\cite{ahmadzadeh2006multi} proposed a method to coordinate a team of fixed-wing robots to monitor an area of interest. The main limiting factor for this work is the dependency on a fixed-wing aerial vehicle model. 
Correspondingly, the paths of each robot are restricted to a smaller allowable set of movements. Additionally, consideration for balancing the workload between the robots is not considered, resulting in an uneven distribution of workload across the swarm, leading to an increase in the overall time required to complete the mission.

Other methods, such as~\cite{seyedi2019persistent, maini2020visibility, bailon2018planning, maini2018persistent}, provide solutions to a similar application - persistent monitoring. While persistent monitoring is similar, it generally aims to create routes for robots which will be repeated constantly. With this objective in mind, the areas assigned to each robot are often similar in size, and the travel distance from the initial launch site is not accounted for during assignment. Ultimately, these methods cannot be utilized as viable solutions to the time sensitive coverage problem proposed here, as a lack of attention to the initial states of the robots will result in an uneven distribution of the workload and subsequent increase in total time taken to complete the surveillance. 

Common safety practices have also been implemented as the growth of popularity in UAV usage has surged. New practices such as geofencing, which designate specific "no-fly" zones to restrict aerial vehicles from causing unintended damage to nearby objects and humans~\cite{hermand2018constrained} have been implemented. Previous task planning and coverage methods such as \cite{ahmadzadeh2006multi, danoy2015connectivity, cheng2009distributed} do not present the ability to account for such discontinuities in the environment. 





\section{Scalable Coverage Path Planning Algorithm}
\label{section:methodology}
\vspace{-0.2cm}
Our proposed algorithm - \textbf{\underline{S}}calable \textbf{\underline{Co}}verage \textbf{\underline{P}}ath \textbf{\underline{P}}lanning (\ourMethod) -- is a scalable method, presenting computational efficiencies that is tractable for a large number of robots and/or a large map environment. The method is also designed to provide a load-balanced path plan to minimize the total mission completion time, and works best with multi-rotor type aerial vehicles due to the way in which these paths are constructed. The algorithm takes a user-specified number of robots and boundary points specifying the area which needs to be monitored in geographical coordinates, and returns a time optimal list of way-points for each robot. The algorithm proposed here aims to minimize time required to monitor the whole area of interest, which we refer to as the completion time. To accomplish this, expenditures from travelling during the course of each robot's path is taken into account during the allocation of area per robot. For each robot, this includes the cost of travelling from its initial state to its first position in the area it was assigned, as well as the cost of scouting its assigned area. The battery life of each robot is assumed to be sufficient to complete the mission without the need to recharge or swap the battery. The problem is formulated as a min-max problem aimed at minimizing the worst completion time across all robots; i.e., $f = \min_{r}\max_{r}\,(t_r)$, where $t_r$ describes the total time taken to complete the mission for each robot, $r\in\mathcal{R}=[1..N_r]$, and $N_r$ is the total number of robots.
\begin{figure*}[!ht]
    \centering
    \includegraphics[width=.85\textwidth]{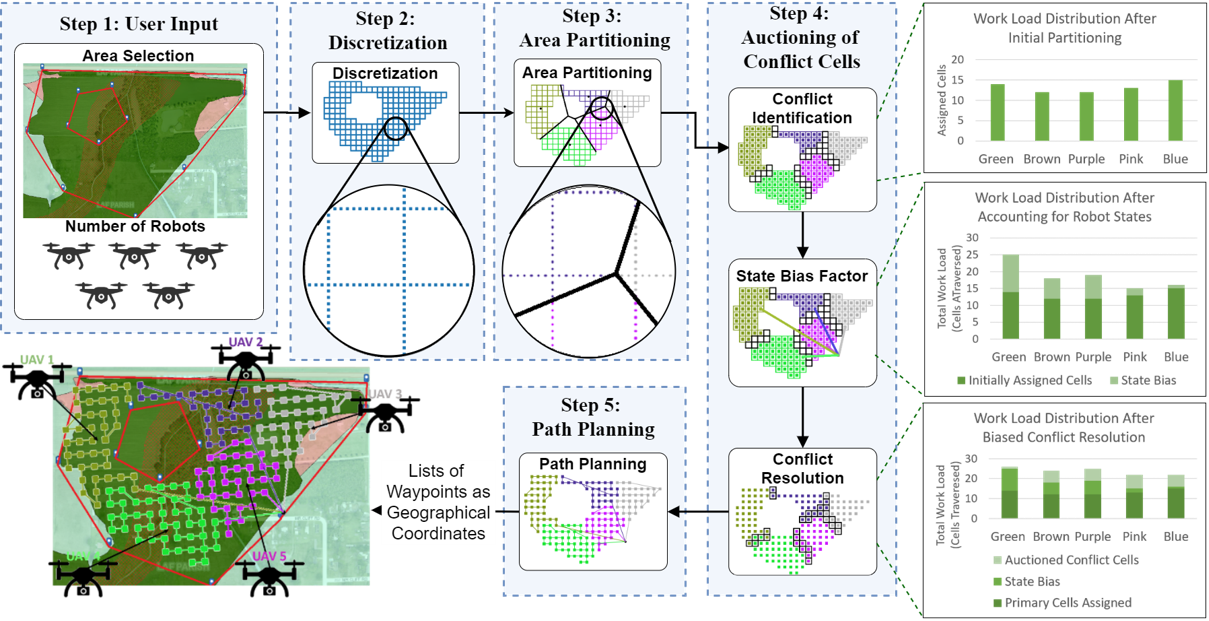}
    \vspace{-9pt}
\caption{Flowchart of the \ourMethod{} algorithm with visualization of each step} \label{fig:QLB-overview}
    \vspace{-15pt}
\end{figure*}
%

Our approach to solve this problem is to perform iterative clustering to create partitions of equal mass from the area of interest after discretization of said area. For simplicity, we assume that the robots maintain constant velocity throughout the duration of their mission ($t_r = \sum_{k=1}^{n_k}\frac{d_k}{v}$). Here, $n_k$ is the total number of cells assigned to each robot, $d_k$ is the distance between each subsequent cell robot $r$ must visit in their assigned list of way-points, and $v$ is the robots' velocity, which is assumed to be constant throughout its mission. Finally, the total surveyed area of the $r^\text{th}$~robot ($A_r$) is computed by $A_r=n_k (W_k)^2$, where $W_k$ is the width of each square cell. The width of the cells is calculated as 
    $W_k = 2h \, \tan{(\frac{F}{2})}\label{eq:cell width}$.
Here, $h$ is the height above ground at which the robots will fly, and $F$ is the downward field of view (FoV) of the robot's onboard camera system. This dependency on the robot's FoV results in square cell sizes, where the width of the cells is a function of the robot's altitude and FoV. With this formulation of the width, we can assume each cell will be surveyed when a robot reaches the center of the cell.

As shown in Fig.~\ref{fig:QLB-overview}, the overall algorithm involves five major steps: \textbf{Step 1 - Area Selection:} The user provides the algorithm with the number of robots available, their initial positions, and a series of points in geographical coordinates (latitude, longitude) which designate the vertices of the polygon bounding the area of interest. \textbf{Step 2 - Discretization:} The algorithm converts this polygon into a pseudo-discrete space, where the perimeter of each cell in the grid space is defined as a series of evenly spaced points. \textbf{Step 3 - Area Partitioning:} The points are then grouped with the help of Lloyd's algorithm - an iterative voronoi-based clustering method, where each point is temporarily assigned to a cluster representing a robot - regardless of their parent cell. \textbf{Step 4 - Auctioning of Conflict Cells:} The previous step results in various "conflicted cells" which do not belong solely to one cluster. These conflicted cells are auctioned off using a bias factor based on the distance between the robots' initial position and their closest currently assigned cell, leading to a more evenly distributed workload amongst the robots. 
\textbf{Step 5 - Path Planning:} The final step, "Path Planning", takes the cells and initial positions assigned to each robot, and creates an time-optimal list of waypoints for the robot to traverse during its mission.

\subsection{Step 1: User Input}
The user need only provide the algorithm with the number of robots available, their initial positions, and a series of points in geographical coordinates (latitude, longitude) which designate the vertices of the polygon bounding the area of interest. This information is transformed into a Cartesian coordinate space with a unit vector equivalent to one meter using the Haversine formula below:
\begin{align}
a &= \sin^2(\Delta\phi/2) + \cos \phi_1 \cos(\phi_2) \sin^2(\Delta\lambda/2)\\ 
d &= 2\;R\;\text{arc}\tan(\sqrt{a}, \sqrt{(1-a)})
\end{align}
where $\phi$ is latitude, $\lambda$ is longitude, $R$ is earth’s radius (mean radius = 6,371km), and d is the distance between the two geographical points.

\subsection{Step 2: Mission Area Discretization}


The robot state information provided by the user is converted from geographical coordinates, to an absolute and continuous coordinate system, where the unit vector is equivalent to one meter. The continuous polygon describing the area of interest is converted into a pseudo-discretized grid space, where the perimeter of each cell in  the  grid  space  is  defined  as  a  series  of  evenly  spaced points, and each cell represents a square area to be ultimately assigned to one robot. The cells are of equal width, and their size is determined by calculating the amount of area which can be seen by an robot at an instant in time, which can found using the equation to compute $W_k$ from earlier. The FoV of the robots essentially determines the cell width, as this is the maximum amount of area each robot will be able to view once positioned in the center of a cell at the specified operating height. 

\subsection{Step 3: Load-Balanced Initial Partitioning}


Clustering is used to create $N$ clusters, where the $x$ and $y$ coordinates of each point in space are used as the only features, using an iteratively updating centroid based on Lloyd's algorithm~\cite{kasim2017quantitative}, with the goal of generating clusters of equal mass, from the points in the grid space.
\begin{equation}
\label{eq:minibatch kmeans}
\min \sum_{x\in X} ||f(C, x) - x||^2
\end{equation}
This equation (Eq.~\eqref{eq:minibatch kmeans}) describes the objective function of the variation of Lloyd's algorithm used here. $f(C, x)$ returns the nearest cluster center $c \in C$ to $x$
using Euclidean distance. The number of clusters, $N_r$ is equal to the number of robots, the convergence tolerance is equal to the cell width, $W_k$ divided by eight, and the maximum number of iterations for the clustering method is set to ten. The resulting clusters are equally sized, and resemble areas created via Voronoi Tessellation with equally sized regions. 


\subsection{Step 4: Auctioning of Conflict Cells}


The continuity of the tessellation boundaries results in cells whose perimeter points belong to more than one cluster group along the boundary regions. Using the states of the robots, these conflicting cells are auctioned off between the robots to balance the work load among them. The algorithm calculates the minimum distance from the closest cell in it's assigned area and uses this to add a state bias factor per robot, which is scaled according to this distance. The calculation of this factor is expressed in Eq.~\eqref{eq:distance bias}. The robots' parameters such as FoV, operational height, and velocity are used to approximate the time required for each robot to complete surveillance of its assigned area. Algorithm~\ref{algorithm:resolve conflicts} illustrates the flow of this process. The end result is $N$ lists containing the locations of cells assigned to each robot.
\begin{equation}
    \label{eq:distance bias}
    d_B(r) = d_0(r) \times B
\end{equation}
This equation describes the distance bias, $d_B(r)$, added to each robot's workload total when auctioning off conflicting cells. The resulting workload for each robot is evaluated when redistributing conflicting cells. Here, $d_0(r)$ is the distance robot $r$ must travel to arrive at its first cell from its initial position, and $B$ is a heuristic bias factor which determines the level of influence $d_0(r)$ has on workload redistribution. More specifically, increasing $B$ decreases the amount of workload assigned to robots with higher initial travel distances. For the experiments in this paper, a setting of $B=0.5$ was observed to provide adequate performance. 

\begin{algorithm}[b]
\KwResult{Unique list of cells per robot}
C = Conflicting cell list \;
\For{Cell in C}{
Lowest cell count = $\inf$ \\
Lowest cell count robot = None \\
\For{Robot in Swarm}{
    \eIf{{Robot's cell count + Robot Bias ($d_B(r)$)} < Lowest cell count}{
    Lowest cell count = Robot's cell count
    Lowest cell count robot = Robot
    }{
    continue
    }}
Add $Cell$ to Lowest cell count robot's cell list
}
\caption{Resolution of Conflict Cells}\label{algorithm:resolve conflicts}
\end{algorithm}

\begin{algorithm}[b]
\KwResult{Optimized list of way-points per robot}
\For{$Robot$, $Cells$}{
    $Position$ = Robot initial position \\ 
    $Task list$ = Robot cells \\ 
    $Optimal path$ = list \\
    \While{Length($Task list$) > 1}{
        Add $Position$ to $Optimal path$ \\ 
        Run NN on Task list \\ 
        NN = NN($Position$) \\ 
        Remove $Position$ from $Task list$ \\ 
        $Position$ = NN \\ 
    }
    Add $Task list$ to $Cells$
}
\caption{Discrete Path Planning}\label{algorithm:path planning}
\end{algorithm}

\subsection{Step 5: Path Planning}
Once the robots' areas have been assigned, each list of cells assigned to the robots is reordered into a list of optimal way-points which minimizes the energy consumption of the given robot when completing its assigned set of cells. The planning problem here is modelled as a nearest neighbor problem, where the next optimal locations would be the cell closest to the current position, and is solved using a KD-tree algorithm~\cite{bentley1975multidimensional} (leaf size of 10) for increasing computational efficiency. This proves to be an efficient and reliable means of planning given that the areas are discrete and convex. For a given robot, a list containing just the starting position of the robot is created, along with a copy of the list of all cell locations the robot must visit, including its starting position. For each iteration of the path planning algorithm, the cell nearest to the robot's current, or $i^{th}$, cell position is found using the KD-tree algorithm. This nearest cell is then added to the optimal way-point list, while the $i^{th}$ cell is removed from the list of possible nearest neighbors. The algorithm iterates in this manner until the list of cells which need to be visited is empty, resulting in a list of optimal way-points for the robot. This algorithm is illustrated in Alg.~\ref{algorithm:path planning}. Finally, the waypoints are transformed back into geographical coordinates using several matrix transformations with the following equation.
\begin{equation}
\label{eq:matrix transformations}
    P_g = P_cC^{-1} G
\end{equation}
Here, $P_g$ is the point(s) to be transformed to the geographical plane, conversely, $P_c$ is the point(s) to be transformed from geographical plane. C is the Cartesian transformation matrix, and $G$ is a vector containing information about the bounds in geographical coordinates.

\section{Case Study: Multi-UAV Post-Flood Assessment}\label{sec:case study}
In this section, We demonstrate the utility of our \ourMethod{} algorithm on post-flood assessment performed by multiple UAVs. Aerial imagery can assist in flood disaster response efforts, as they contain crucial, recent spatial information of the area \cite{adams2011survey, vetrivel2018disaster}. Flood disaster situations are difficult to traverse, require knowledge of the environment as quickly as possible, and may include areas who's information can not, or may not need to be collected - making this an ideal test case for our multi-robot coverage path planning algorithm. Below we provide a detailed explanation of the design of experiments, as well as the simulation and algorithm settings. 

\subsection{Design of Experiments}
\begin{figure}[!t]
    \centering
    \subfigure[Small map area]
    {\includegraphics[height=1.2in]{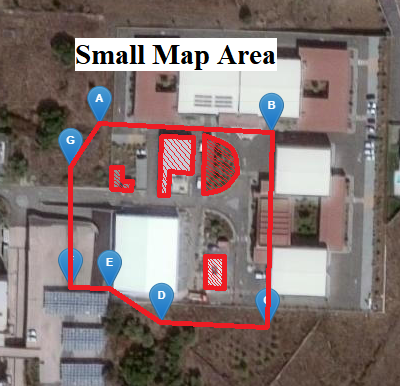}\label{fig:smallMap}}%
    \hspace{0.1cm}
    \subfigure[Medium and Large map area]
    {\includegraphics[height=1.2in]{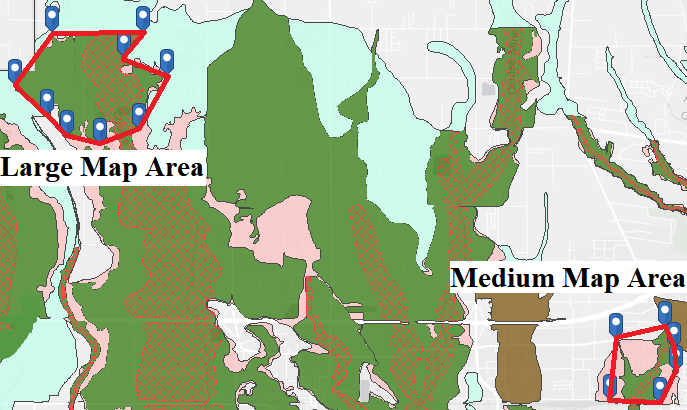}\label{fig:medLargeMap}}
    \vspace{-4pt}
\caption{Three test areas (small, medium, and large maps) that are used in this study. (a) Small map for comparative analysis used in Guastella et al. \cite{guastella2019coverage}. The red regions are no-fly zones. (b) A flood map of Lafayette, Louisiana provided by FEMA. The green areas highlight the highest risk flood areas in the region. The markers indicate the points of the polygon bounding the flood regions to be surveyed using UAVs with \ourMethod.} \label{fig:full_lafayette_map}
\vspace{-15pt}
\end{figure}

\noindent\textbf{Selection of Geography}: 
The city of Lafayette, Louisiana was chosen for this test given the area's track record of severe floods. A publicly available FEMA flood map (shown in Figure \ref{fig:full_lafayette_map}) was used to identify the highest risk areas for flooding. Three areas in Lafayette, Louisiana were chosen to represent medium (1.012 \sqkm) and large (3.436 \sqkm) flood areas to validate the effectiveness of the algorithm at varying map sizes. 



\noindent\textbf{UAV Characteristics:} The Federal Aviation Administration (FAA) states that the maximum allowable altitude for a small unmanned aircraft is 400 feet (121.92 meters) above the ground, higher if your drone remains within 400 feet of a structure. Additionally, the maximum allowable speed for the same aircraft is 100 mph (44.704 m/s). Based on these limitations, and commercially available unmanned aircraft specifications, we set the lower and upper bounds of the average speed at 2 m/s and 10 m/s, respectively. Similarly, based on these sources, we choose the lower bound for operating height to be 25 meters. However, due to the limitations of the cameras capturing data, the upper bound for the operating height is chosen to be 100 meters, based on the minimum height required to identify humans, according to industry standard Detection, Recognition, and Identification (DRI) ratings and requirements \cite{optics_2017}. As the height of the camera increases with the aerial vehicle, it can cover larger area. However, this is at the expense of a loss in detail as less pixels in the image are used to illustrate an area. The minimum image size required to recognize a human is 13 by 5 pixels \cite{optics_2017}. Using this information, and a  (FoV) calculator, we can find the maximum height a camera could be used for an aerial victim search application. A dual camera system, with capability of capturing both thermal and RGB images, is assumed. Thermal image resolution is typically lower than RGB images, and is the limiting factor in obtaining valid optical imagery. For this study, we assume a maximum resolution of $640\times512$ with a focal length of 35 millimeters and a FoV of $18\times14$ degrees. 
\subsection{Simulation and Algorithm Settings}
We implemented \ourMethod{}~\footnote{{The source code is now publicly available at \url{https://github.com/adamslab-ub/SCoPP}}} in Python using the sci-kit~\cite{scikit-learn} library for iterative clustering, and mlrose~\cite{Hayes19} and networkx~\cite{hagberg2008exploring} libraries for graph operations. All simulations were run on a desktop with Intel i7 CPU with 4 cores at 4 GHz and 32GB RAM. 
The settings of \ourMethod{} are set as follows. \noindent\textbf{Initial Partitioning:} 
The number of clusters is set to be equal to the number of robots. The tolerance criteria has been set to one eighth the width of the cell (i.e., $W_k/8$). 
\noindent\textbf{Path Planning:} 
The KD-tree setting was used due to its computational efficiency in comparison with brute force at larger sample sizes, $N$, with dimension, $D=2$. Brute force scales as $O[DN^2]$ while KD-tree scales as $O[DN log(N)]$ or better \cite{scikit-learn}. For larger map sizes, fewer robots, and/or smaller cell sizes, $N$ can grow to be well within the hundreds. Since we choose the immediate nearest neighbor, we chose to search for a nearest neighbor size of 2.

\section{Results and Discussion}\label{sec:experiments}
\label{section:results}
To evaluate and study the effectiveness of our proposed method, we conduct three different analyses. First, we compare our primary method with another state-of-the-art method (\textit{Comparative Analysis}). Second, we study how the performance of the swarm varies across different swarm size (\textit{Scalability Analysis}). Finally, we study how the computational time of our primary method is affected by varying swarm sizes and maps (\textit{Computing Time Analysis}).

\subsection{Comparative Analysis of \ourMethod{}}
Table~\ref{tbl:compareMethods} summarizes the comparison of our \ourMethod{} method to one state-of-the-art method in terms of mission time.
The comparative method, which we refer to as Guastella's method, uses a Morse-based exact decomposition \cite{acar2002morse}, and utilizes a back-and-forth survey method per cell \cite{guastella2019coverage}, which has been successfully used for multi-robot coverage path planning. The source code of this baseline method was shared with us by the original authors on request, thereby further motivating this comparison. To generate results for our comparative analysis, we use this original code, and its given default settings (i.e., 13-robots and studied map). Both methods are run on the small map (Fig.~\ref{fig:smallMap}). To create a fair comparison, our method is run with the same settings as the reference material; i.e.,  robot velocity of 4 $m/s$. Our method is stochastic, due to partitioning and path planning (Steps 3 \& 5). Due to this, \ourMethod{} is run 100 times to report its performance in terms of (mean $\pm$ std-dev of) mission completion time.
Table~\ref{tbl:compareMethods} shows that \ourMethod{} outperforms Guastella's method by providing a mission time that is 18 times faster than Guastella's. We are not able to compare the computational cost of each method, since the provided code for Guastella is in Matlab and our code is in Python.
\begin{table}[!h]
\centering
\caption{Comparative analysis of \ourMethod{} with a team of 13 robots.} 
\label{tbl:compareMethods}
\vspace{-10pt}
\begin{tabular}{|l|c|}
\toprule
\textbf{Method}  &  \textbf{Completion Time (s)}\\
\midrule
 \ourMethod{} (Ours) & $196.53 \pm 14.53$\\
 Guastella's & $3,591.8 \pm 0.00$\\
 \bottomrule
\end{tabular}
\vspace{-5pt}
\end{table}

\subsection{Scalability Analysis of \ourMethod{}}
\begin{figure}
    \centering
    \includegraphics[width=.9\columnwidth]{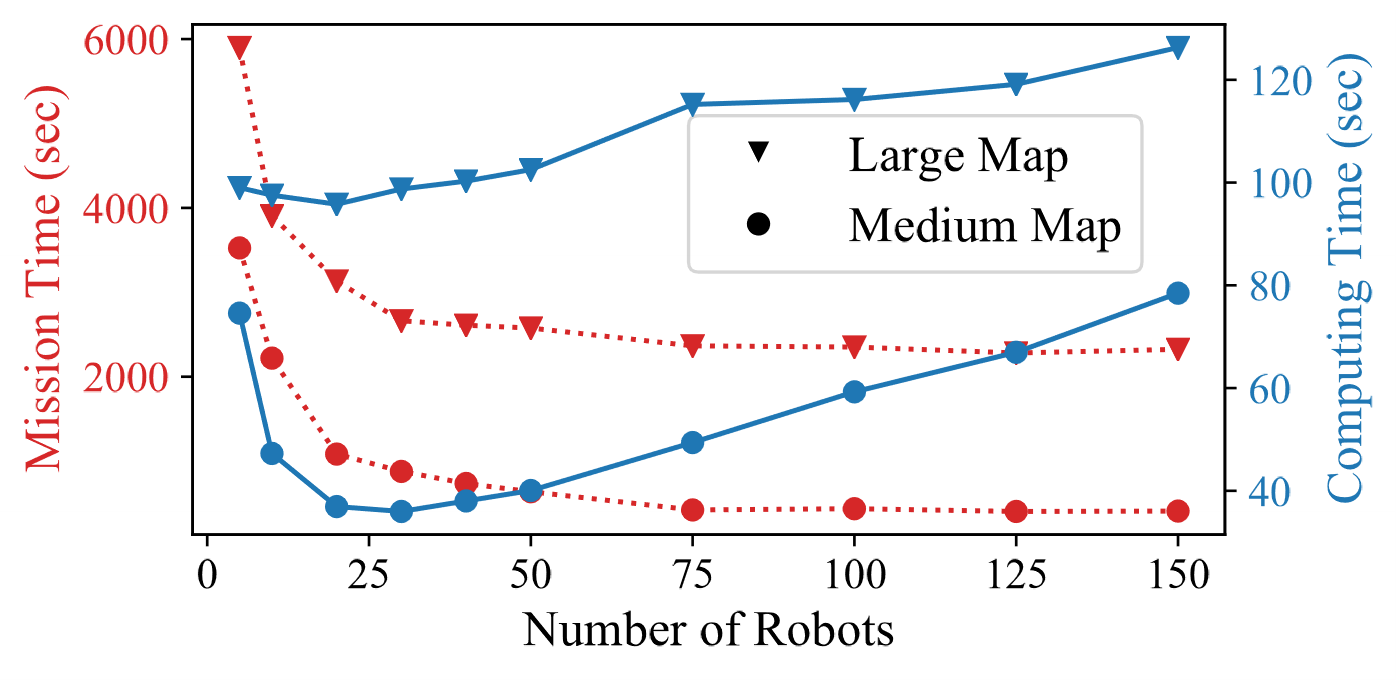}
    \vspace{-10pt}
    \caption{Scalability Analysis: Mission time and computing time of \ourMethod{} over different swarm sizes in two medium and large maps.} 
    \label{fig:scalability_analysis}
    \vspace{-15pt}
\end{figure}

In order to study how the performance of the \ourMethod{} algorithm in terms of both mission time and computing time changes by varying the number of robots, we run the algorithm on the medium and large maps (Fig.~\ref{fig:medLargeMap}) with different number of robots, varying from 5 to 150. Figure~\ref{fig:scalability_analysis} shows the obtained result for this experiment. As it can be seen from Fig. \ref{fig:scalability_analysis}, for both maps the mission time decreases with increasing number of robots, and saturates after a certain point, i.e., after $75$, due to the decreasing marginal utility of additional team members. In the medium and large maps, the mission time drops respectively about 75.09\% and 54.86\% as the number of robots in the team increases from 5 to 30. 

In terms of computing time, we observe that the computing time initially decreases with increasing the number of robots, but at certain point it starts increasing; i.e., at $N_r=30$ and $N_r=20$ in the medium and large maps, respectively. The decreasing trend for small number of robots is more extreme and it will be discussed in Section~\ref{ssec:computational_cost}. It should be noted that the worst computing time of \ourMethod{} is 80 seconds and 120 seconds in the medium map the computing time, respectively. This clearly shows the applicability of the proposed algorithm for planning for a large team of robots. 

\subsection{Computing Time Analysis}\label{ssec:computational_cost}
\begin{figure}
    \centering
    \includegraphics[width=1\columnwidth]{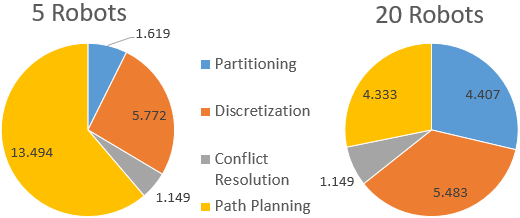}
    \caption{Computation time profiling between various phases of \ourMethod{} on the medium map size.}
    \label{fig:computing_cost}
    \vspace{-15pt}
\end{figure}

To explain the above observation (i.e., the  computing time initially decreases with increasing the number of robots from 5 to 20), we provide a time profiling of \ourMethod{} for two swarm size 5 and 20 for the medium map. Figure~\ref{fig:computing_cost} shows the computing time of each step in \ourMethod, namely: i) Discretization (Step 2), ii) Partitioning (Step 3), iii) Conflict Resolution (Step 4), and iv) Path Planning (Step 5). The main observations from Fig.~\ref{fig:computing_cost} can be summarized as follows:
\begin{enumerate}
    \item \textit{Conflict Resolution} is the least expensive step and remains the same value for both 5- and 20-robot cases.
    \item The computing time of \textit{Partitioning} increases linearly with increasing the number of robots. This is expected since the average computational complexity of the Lloyd's algorithm is linear, its worst-case is super-polynomial.
    \item The computing time of the \textit{Discretization} step is remains approximately the same. This is due to this fact that the number of robots is invariant to  discretization.
    \item \textit{Path Planning} forms the major cost of the 5-robot case and it decreases $\sim$68\% for the 20-robot case. The computational complexity of path planning is proportional to the number of nodes (here cells) for each partition (i.e., $n\log^2 n$, where $n$ is the number of nodes). With smaller swarm size, each partition is larger and contains more cells to be visited by a robot.
\end{enumerate}
The above observations and Fig.~\ref{fig:scalability_analysis} show that there is a sweet spot that minimizes the overall computing time of \ourMethod; here, the sweet spot occurs at $N_r=20$. For larger swarm size, the computing time of \textit{path planning} decreases, the computing time of \textit{conflict resolution} and \textit{discretization} remains approximately the same, but the computing time of \textit{partitioning} keeps growing. It can be concluded that the partitioning step is the main bottleneck for the scalablity, in terms of increasing swarm size, of \ourMethod.

Increasing the size of the area of interest is another key factor that impacts the computing time of all four steps of \ourMethod. Figure~\ref{fig:scalability_analysis} illustrates that the overall computing time of \ourMethod{} on the large map (3.44 \sqkm) can be anywhere from 24.67\% to 63.56\% higher in comparison with the medium map (1.01 \sqkm).

\section{Conclusion} \label{section:future work conclusions}
\vspace{-0.1cm}
This study provided a novel multi-robot coverage path planning algorithm (\ourMethod{}) for optimizing the load balance between robots by taking into account their initial states, while minimizing the completion time for monitoring non-convex areas. The method uses an iterative clustering algorithm to create zones for each robot over a continuous space while accounting for geofenced zones and the robots' initial states to balance the load each robot undertakes. The \ourMethod{} method is observed to reduce the mission completion time by an average of 94.53\% in comparison with a recent state-of-the-art method. With regards to scalability, the results show a promising linear relationship between computing time and team size, e.g., the computing time merely doubles when team size increases 6-fold on the medium map. This makes \ourMethod{} uniquely useful for large swarm-robotic or multi-UAV applications. We also show that \ourMethod{} is highly scalable w.r.t. the map area to be covered, e.g., the computing cost increased by only 63.56\% when going from the medium map (1.01 \sqkm) to the large map (3.44 \sqkm).

The proposed algorithm currently applies to homogeneous swarms or teams with different initial states. However, in certain applications, one may require several types of robots with varying capabilities to monitor an area. One of the immediate future steps is further adaptation of the algorithm to create plans for a team of heterogeneous robots. Furthermore, the battery life for all robots in this study was assumed to be sufficient for the entire mission. In practice, this may not always be the case. The algorithm can then be modified such that when the robots are running low on battery, they are allowed to return to depot, which along with future field implementations will help further validate the competitive advantages of this scalable coverage planning approach. 


\bibliographystyle{unsrt}
\bibliography{references,reference-ghassemi}





\end{document}